\documentclass[runningheads]{llncs}
\usepackage[dvipdfmx]{graphicx}

\usepackage{amsmath, amssymb}

\DeclareMathOperator*{\argmin}{argmin}
\DeclareMathOperator*{\argmax}{argmax}

\usepackage{booktabs}
\usepackage{color}

\begin{document}
\title{Balancing Cost and Quality: An Exploration of Human-in-the-loop Frameworks for Automated Short Answer Scoring}

\titlerunning{Balancing Cost and Quality: Frameworks for ASAS}

\author{Hiroaki Funayama\inst{1,2}, Tasuku Sato\inst{1,2}, Yuichiroh Matsubayashi\inst{1,2},  Tomoya Mizumoto\inst{3,2}, Jun Suzuki\inst{1,2} and Kentaro Inui\inst{1,2}
}
\authorrunning{Funayama et al.}

\institute{
 Tohoku University, Sendai, Japan\\
 \email{\{h.funa, 	tasuku.sato.p6\}@dc.tohoku.ac.jp,} \email{\{y.m, jun.suzuki,	inui\}@tohoku.ac.jp,}
 \and
 RIKEN, Tokyo, Japan
 \and
 Future Corporation, Tokyo Japan\\
 \email{t.mizumoto.yb@future.co.jp}
 }

\maketitle              
\begin{abstract}

Short answer scoring (SAS) is the task of grading short text written by a learner.
In recent years, deep-learning-based approaches have substantially improved the performance of SAS models, but how to guarantee high-quality predictions still remains a critical issue when applying such models to the education field.
Towards guaranteeing high-quality predictions, we present the first study of exploring the use of \emph{human-in-the-loop} framework for minimizing the grading cost while guaranteeing the grading quality by allowing a SAS model to share the grading task with a human grader.
Specifically, by introducing a confidence estimation method for indicating the reliability of the model predictions, one can guarantee the scoring quality by utilizing only predictions with high reliability for the scoring results and casting predictions with low reliability to human graders.
In our experiments, we investigate the feasibility of the proposed framework using multiple confidence estimation methods and multiple SAS datasets. We find that our human-in-the-loop framework allows automatic scoring models and human graders to achieve the target scoring quality.
\end{abstract}
%
%
%

\section{Introduction}

\begin{figure}[t]
\centering
\includegraphics[width=\textwidth]{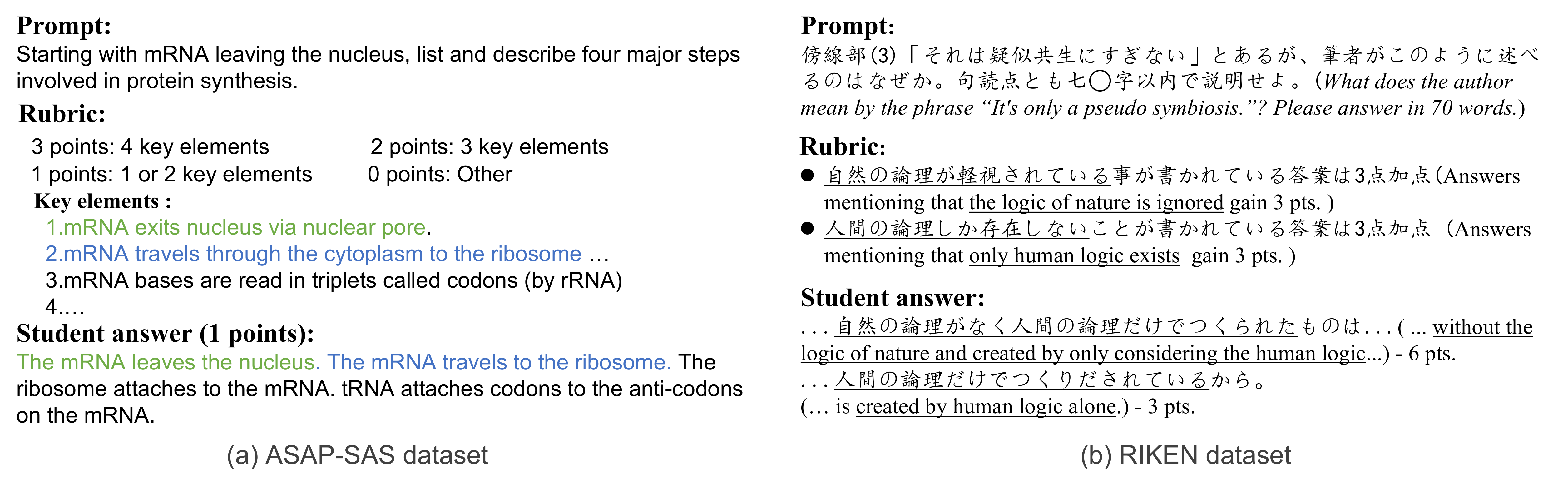}
\caption{(a) Examples of rubrics and student's answers of prompt 5 excerpted from kaggle Automated Student Assessment Prize Short Answer Scoring (ASAP-SAS) dataset, and (b) example of a prompt, scoring rubric, and student's answers excerpted from RIKEN dataset\cite{mizumoto-etal-2019-analytic}, translated from Japanese to English. For space reasons, some parts of the rubrics and answers are omitted. }
\label{fig:rubrics_and_answer}
\end{figure}

Short answer scoring (SAS) is a task used to evaluate a short text written as input by a learner based on grading criteria for each question (henceforth, prompt).  Figure~\ref{fig:rubrics_and_answer} gives examples of rubrics and student's answers. Automatic SAS systems have attracted considerable attention owing to their abilities to provide fair and low-cost grading in large-scale examinations and to support learning in educational settings~\cite{Singla2019GetIS,Leacock}. In recent years, deep -learning-based approaches have improved the performance of automated scoring models~\cite{riordan-etal-2017-investigating}. However, the possibility of errors produced by an automatic grading system cannot be completely eliminated, and such errors may interfere with the learner's learning process~\cite{sychev2020automatic}. Owing to a concern for such automatic scoring errors, current automatic grading systems are often used as references for human graders to detect grading errors~\cite{Attali_Burstein_2006}. To further utilize automatic scoring systems in the education field, it is critical to guarantee high-quality grading.

To tackle this challenge, we propose a human-in-the-loop automatic scoring framework in which a human grader and an automatic scoring system share the grading in order to minimize the grading cost while guaranteeing the grading quality.
In this framework,  we attempt to guarantee the scoring quality using confidence estimation methods by adopting only the highly reliable answers among the automatic scoring results and cast the remaining answers to human graders (see Fig. \ref{fig:overall}).  
Specifically, we perform the following two-step procedure: (1) estimating the threshold of confidence scores to achieve the desired scoring quality in the development set, and (2) filtering the test set using the determined threshold. 
The achievement of the entire framework is evaluated in two aspects: the percentage of automatically scored answers and how close the scoring quality is to the predetermined scoring quality.

In the experiments, we simulated the feasibility of the proposed framework using two types of dataset with a general scoring model and several confidence estimation methods. 
The experimental result verifies the feasibility of the framework. 
The contributions of our study are as follows. 
(i) We are the first, to the best of our knowledge, to provide a realistic framework for minimizing scoring costs while aiming to ensure scoring quality in automated SAS. 
(ii)  We validated the feasibility of our framework through cross-lingual experiments with a general scoring model and multiple confidence estimation methods. 
(iii) We gained promising results showing that our framework enabled the control of the scoring quality while minimizing the human grading load, and the framework worked well even for prompts for which the agreement between human graders is relatively low. 
The code for our experiments and all experimental setting information is publicly available.\footnote{ \url{https://github.com/hiro819/HITL\_framework\_for\_ASAS}}


\section{Previous research}
In recent years, SAS has attracted considerable attention owing to its ability to provide a fair and low-cost scoring in large-scale exams~\cite{Burrows2015}.
A central challenge in the use of SAS has focused primarily on improving the performance of scoring models~\cite{Williamson_2012}. 
With the recent advancement of models using deep learning, the performance of scoring models has also significantly improved~\cite{riordan-etal-2017-investigating,Singla2019GetIS}.
Towards realizing the practical use of SAS systems in the real world, several researchers have explored outputting useful feedback for an input response~\cite{Woods_2017}, utilizing rubrics for scoring~\cite{wang-etal-2019-inject}, and investigating adversarial input in SAS~\cite{ding-etal-2020-dont}, to name a few. In addition, research on various languages has also been reported in recent years, including Indonesian~\cite{Herwanto-ukara2018}, Korean~\cite{Jang-kass-2014}, and Japanese~\cite{mizumoto-etal-2019-analytic}. 

To the best of our knowledge, the only study in which the use of confidence estimation in SAS was investigated by Funayama et al~\cite{funayama-etal-2020-preventing}.
They introduced the concept of unacceptable critical scoring errors (CSEs).
Subsequently, they proposed a new task formulation and its evaluation for SAS, in which automatic scoring is performed by filtering out unreliable predictions using confidence scores to eliminate CSEs as much as possible.

In this study, we extend the work of Funayma et al~\cite{funayama-etal-2020-preventing}, and propose a new framework for minimizing human scoring costs while controlling the overall scoring quality of the combining human scoring and automated scoring.
We also conducted cross-lingual experiments using a Japanese SAS dataset, as well as the ASAP dataset commonly used in the SAS field.
The human-in-the-loop approach in SAS was also used in a previous work to apply active learning to the task~\cite{horbach-palmer-2016-investigating}; however, our study is unique in that we have attempted to minimize scoring costs while focusing on ensuring overall scoring quality.

\section{Short Answer Scoring: Preliminaries}
\subsection{Task definition}

Suppose $\mathcal{X}$ represents a set of all possible student's answers of a given prompt, and ${\mathbf{x}} \in \mathcal{X}$ is an answer. 
The prompt has an integer score range from 0 to $N$, which is basically defined in rubrics.
Namely, the score for each answer is selected from one of the integer set $S =  \{0, ..., N\}$.
Therefore, we can define the SAS task as assigning one of the scores $s \in S$ for each given input ${\mathbf{x}} \in \mathcal{X}$.
Moreover, to construct an automatic SAS model means to construct a mapping function $m$ from every input of student answer ${\mathbf{x}} \in \mathcal{X}$ to a score $s \in S$, that is, $m : \mathcal{X} \rightarrow S$.

\subsection{Scoring model}
\label{sec:scoring}
A typical, recent approach to constructing a mapping function $m$ is the use of newly developed deep neural networks (DNNs).
As discussed a priori, the set of scores $S$ consists of several consecutive integers $\{0, \dots, N\}$.
We often consider each discrete number as one class so that the SAS task can be treated as a simple $N+1$-class classification task.
This means that each integer in $S$ is considered as a class label, not a consecutive integer.
In this case, a SAS model is often constructed as a probabilistic model.
Suppose $\mathcal{D}$ is training data that consist of a set of actually obtained student's answers $\mathbf{x}$ and its corresponding human annotated score $s$ pairs, that is, $\mathcal{D}=( (\mathbf{x}_i, s_i) )^{I}_{i=1} $, where $I$ is the number of training data.
To train the model $m$, we try to minimize the empirical loss 
on training data $L_{m}(\mathcal{D})$ that consist of the sum of negative log-likelihood for each training data calculated using model $m$. 
Therefore, we can write the training process of the SAS model as the following minimization problem:
\begin{align}
{m} = \argmin_{m^{\prime}} \left\{ L_{m^{\prime}}(\mathcal{D}) \right\}, \quad 
L_{m}(\mathcal{D})  =
- \sum_{(\mathbf{x},s)\in \mathcal{D}} \log \left( p_{m}(s\mid\mathbf{x}) \right)
,
\end{align}
where $p_{m}(s\mid\mathbf{x})$ represents the probability of class $s$ given input $\mathbf{x}$ calculated using model $m$.
Once ${m}$ is obtained, we can predict the score $\widehat{s}$ of any
input (student answer) by using trained model $m$.
We often use the argmax function for determining the most likely score $\widehat{s}$ given $\mathbf{x}$ as
\begin{align}
\label{eq:argmax}
\widehat{s} = \argmax_{s} \left\{ p_{{m}}(s\mid\mathbf{x}) 
\right\}
.
\end{align}


\section{Proposed Framework}
\label{sec:proposed_framework}

\begin{figure}[t]
\centering
\includegraphics[width=\textwidth]{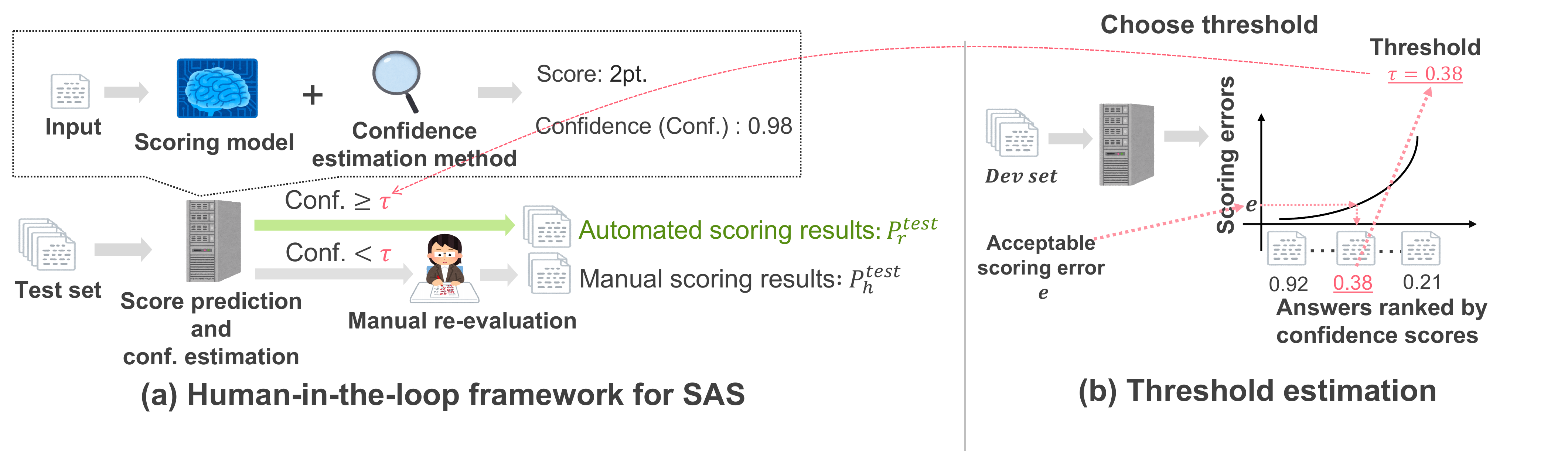}
\caption{Overview of the proposed human-in-the-loop framework for SAS.}
\label{fig:overall}
\end{figure}

Guaranteeing a high-quality scoring framework is crucial, especially when applying it to the education field in actual use.
To address this issue, we propose a novel human-in-the-loop framework that can minimize the grading cost of human graders while guaranteeing the overall grading quality at a certain level.

\subsection{Scoring framework overview}
The basic idea of our scoring framework is intuitive and straightforward.
We first score all the student answers by the method described in Sect. \ref{sec:scoring}, and simultaneously estimate the confidence of the predicted scores.
We then ask trained human graders to reevaluate the student answers when their confidence scores are below the predefined threshold.
Fig.~\ref{fig:overall}(a) shows an overview of our scoring framework.
Our human-in-the-loop framework is based on the assumption that automated SAS systems cannot achieve zero-error, while trained human graders can achieve it.
Therefore, the unreliable scores obtained are reevaluated by human graders to realize high-quality scoring.

The goal of our framework is to minimize the cost of human graders while maintaining erroneous scoring given by automated SAS systems to a minimum.
Thus, the critical components of our framework are the methods o estimating the confidence of scoring and detemining the threshold of reasonable confidence level in advance.

\subsection{Utilizing confidence estimation for SAS}
We define confidence $C$ as a function that is determined using three factors: model $m$, input student answer $\mathbf{x}$, and score $\widehat{s}$ predicted from $m$, that is, $C(\mathbf{x}, \hat{s}, m)$.

Suppose we have $\ell$ unique answers to grade $(\mathbf{x}_1, ... ,\mathbf{x}_\ell)$. 
Let $P$ be the set of all the student answers and its corresponding score pairs.
Let $P_r$ be a subset of $P$, whose confidence is above the predefined threshold $\tau$.
Additionally, $\overline{P}_r$ is the complementary set of $P_{r}$.
Namely, 
\begin{equation}
P=\{(\mathbf{x}_i,\hat{s}_i)\}_{i=1}^{\ell},
\,\,\,
{P}_{r}
= \{ (\textbf{x}, \hat{s}) \in P \mid C(\textbf{x}, \hat{s}, m) \geq \tau
\},
\,\,\,\mbox{and}\,\,\,
\overline{P}_r =P \backslash P_r
.
\end{equation}

We treat $P_{r}$ and $\overline{P}_{r}$ as the reliable and unreliable scoring results, respectively.
In this study, we consider that the answers in $\overline{P}_r$ need rescoring by trained human graders.

The set of pairs of the answers in $\overline{P}_r$ and their corresponding regraded score by human graders is denoted by $P_h$. 
Finally, we combine $P_r$ and $P_h$ as the set of final scoring results, $P_f$.
The relation can be written as $P_h = P_f\backslash {P}_r$.

\subsection{Threshold estimation}
It is difficult to determine the optimal confidence threshold $\tau$ in advance.
To find a reasonable one, we use the development set, a distinct dataset from the training data used for obtaining the model $m$.
Let $P^{\mathtt{dev}}$, $P^{\mathtt{dev}}_r$ and  $P^{\mathtt{dev}}_f$ be 
$P$, $P_r$ and $P_f$, respectively, based on the development data.
For a given, acceptable scoring error $e$, we find the confidence threshold $\tau$ such that the errors of scoring results in the development set do not exceed $e$ \footnote{We assume that the acceptable scoring error is determined by test administrators.}.
\begin{align}
\label{eq:determine_threshold}
\tau  =
\argmax_{\tau^{\prime}}
\left\{|{P^{\mathtt{dev}}_{r^{\prime}}}|\right\}
\quad \text{subject to}  \,\,\,\, \texttt{Err}(P^{\mathtt{dev}}_f) \leq e 
,
\end{align}
where $\texttt{Err}(P^{\mathtt{dev}}_f)$ represents the sum of scoring errors occurring within $P^{\mathtt{dev}}_f$.
\subsection{Evaluation}
${|{P}^{\mathtt{test}}_{r}|} /  {|{P}^{\mathtt{test}}_{f}|}$ is the ratio of automatically scored answers (i.e., automatic scoring coverage), and $\texttt{Err}({P}^{\mathtt{test}}_{f})$ represents the grading error in the set of final answer-grade pairs for the test set ${P}_f^{\mathtt{test}}$. 
In our experiments, we evaluate the performance from both the grading error on the test set and the automatic grading coverage ratio.

\section{Experiments}
In the experiments, we verify the feasibility of the proposed framework using three confidence estimation methods as our test cases.
We publish the code and instance IDs used for training, development and test data.\footnote{\url{https://github.com/hiro819/HITL\_framework\_for\_ASAS}}

\subsection{Base scoring model}
\label{ssec:scoring_model}
We selected the BERT~\cite{devlin-etal-2019-bert} based classifier as the base scoring model because this is one of the most promising approaches for many tasks in text classification.
We can also expect to achieve near state-of-the-art performance in our experiments.

The model first converts the input student answer ${\mathbf{x}}$ into a feature representation $\mathbf{Z}\in\mathbb{R}^{d_h\times n}$ using the BERT encoder $\texttt{enc}(\cdot)$, that is, $\mathbf{Z} = \texttt{enc}(\mathbf{x})$, where $d_h$ is the vector dimension and $n$ is the number of words (tokens) in $\mathbf{x}$.
The model then calculates the probability $p_m(s \mid \textbf{x})$ using the vector of CLS token
$\mathbf{h}^{\rm (CLS)}$, which is the first token in the BERT model, in $\mathbf{Z}$ with the following standard Affine transformation and softmax operation:
\begin{eqnarray}
    \label{prediction_classifier}
    p_m(s\mid\mathbf{x})=&
    {\textbf{o}_{s}}^{\top} \texttt{softmax}
    \left(
    \mathbf{W}\mathbf{h}^{\rm (CLS)} + \mathbf{b}
    \right)
    ,
\end{eqnarray}
where $\mathbf{W}$ and $\mathbf{b}$ are the trainable model parameters of $(N+1)\times d_h$ and $(N+1)\times 1$ matrices, respectively.
$\textbf{o}_s$ is the one-hot vector, where 1 is located at score $s$ in the $d_h$-vector.
Note that $\textbf{o}_s$ is a kind of selector of a score in the vector representation generated from the softmax function.
Finally, we used the standard operation, Eq.~(\ref{eq:argmax}), to obtain the predicted score $\widehat{s}$.


\subsection{Confidence estimation methods}
\subsubsection{Posterior probability.}
The most straightforward way to estimate the confidence score in our classification model is to consider the prediction probability (posterior) calculated Eq. (\ref{prediction_classifier}):

\begin{equation}
    C_{\mathtt{prob}}(\mathbf{x}, \hat{s}, m) = {p_m(\widehat{s}\mid \mathbf{x})}.
\end{equation}

\subsubsection{Trust score.}
Funayama et al.~\cite{funayama-etal-2020-preventing} used the \textit{trust score}~\cite{DBLP:conf/nips/JiangKGG18} to estimate confidence in automatic scoring of written answers. Here, we also consider using the \textit{trust score} as one of the confidence estimation methods. 

We calculate the trust score as follows. The first step is to input the training data $\{({\mathbf x}_1, s_1), ..., ({\mathbf x}_k, s_k)\}$ into the trained automatic scoring model $m$ and obtain a set of feature representations $H=\{{\mathbf{h}}_1, ..., {\mathbf{h}}_k\}$ corresponding to each training instance. 
In addition, the feature representations are kept as clusters $H_s =\{{\mathbf{h}}_i\in{H}\mid s_i = s\}$ for each score label $s$.
We then obtain the feature representation $\mathbf{h}$ and predicted score $\hat{s}$ for an unseen input ${\mathbf x}$ from the test data.
The \textit{trust score} $C_{\mathtt{trust}}({\mathbf x}, \hat{s}, m)$ for the scoring result $({\mathbf x}, \hat{s})$ is calculated as
\begin{equation}
\label{eq:Trust Score}
C_{\mathtt{trust}}({\mathbf x}, \hat{s}, m) = \frac{d_{c}({\mathbf x}, H)}{d_{p}({\mathbf x}, H)+d_{c}({\mathbf x}, H)}
,
\end{equation}
where $d_{p}({\mathbf x}, H) = \min_{{\mathbf{h}^{\prime}}\in{{H}_{\hat{s}}}}{d({\mathbf{h}}, \mathbf{h}^{\prime})}$ and $d_{c}({\mathbf x}, H) = \min_{ {\mathbf{h}^{\prime}}\in{({H}\setminus{H}_{\hat{s}})}}{d({\mathbf{h}}, \mathbf{h}^{\prime})}$,
and $d({\mathbf{h}},{\mathbf{h}^{\prime}})$ denotes the Euclidean distance from ${\mathbf{h}}$ to ${\mathbf{h}^{\prime}}$.
Note that $H$ can be obtained by using $m$ and $\mathbf{x}$.
To normalize the values in the range from $0$ to $1$, $d_c$ is newly added to the denominator.
We note here that the normalization does not change the original order without normalization.

\subsubsection{Gaussian process.}
Previous works have shown that the performance of regression models is superior to that of classification models in SAS since the regression models treat the score as a consecutive integer, not a discrete category~\cite{johan-berggren-etal-2019-regression,riordan-etal-2017-investigating}.

Gaussian Process Regression (GPR)~\cite{Rasmussen2004} is a regression model that can estimate a variance of the predicted score. 
We utilize this value as the confidence of its prediction and use GPR as a representative of regression models that are compatible with our framework.
We use a publicly available GPR implementation~\cite{gardner2021gpytorch} and train the GPR model on the feature representations output by the trained encoder shown in Section~\ref{ssec:scoring_model}.
\subsection{Dataset}
\label{subsec:dataset}
We conduct cross-lingual experiments using the Automated Student Assessment Prize Short Answer Scoring (ASAP-SAS) dataset, which is commonly used in the field of SAS, and the RIKEN dataset, which is the only publicly available SAS dataset in Japanese.

\subsubsection{ASAP-SAS dataset}
The ASAP-SAS dataset is introduced in kaggle's ASAP-SAS contest.\footnote{\url{https://www.kaggle.com/c/asap-sas}} It consists of 10 prompts and answers on academic topics such as biology and science. The answers are graded by two scorers. The experiment was conducted on the basis of the score given by the first scorer according to the rules of kaggle's ASAP contest. In our experiment, we used the official training data and test data. We use 20\% of the training data as development data.

\subsubsection{RIKEN dataset}
We also used the publicly available Japanese SAS dataset\footnote{\url{ https://aip-nlu.gitlab.io/resources/sas-japanese}} provided in \cite{mizumoto-etal-2019-analytic}. The RIKEN dataset consists of 17 prompts in total.
Each prompt has its own rubric, student answers, and corresponding scores.
All contents were collected from examinations conducted by a Japanese education company, Yoyogi Seminar.
In this dataset, the scoring rubric of each prompt consists of three or four elements (referred to as \emph{analytic criteria} in \cite{mizumoto-etal-2019-analytic}), and each answer is manually assessed on the basis of each analytic criterion (i.e., analytic score). 
For example, the scoring rubric shown in Fig.~\ref{fig:rubrics_and_answer} is the analytic criterion excerpted from Item C in prompt Y14\_2-2\_1\_4 and gives an analytic score to each student answer.
Since each analytic score is given independently from each other, we treat the prediction of each individual analytic score for a given input answer as an independent task. 
For our experiment, we divided the data into 250 training instances and 250 test instances. The training data were further divided into five sets, four sets for training (200 instances), and one set for development (50 instances).

\begin{figure}[t]
\centering
\includegraphics[width=\textwidth]{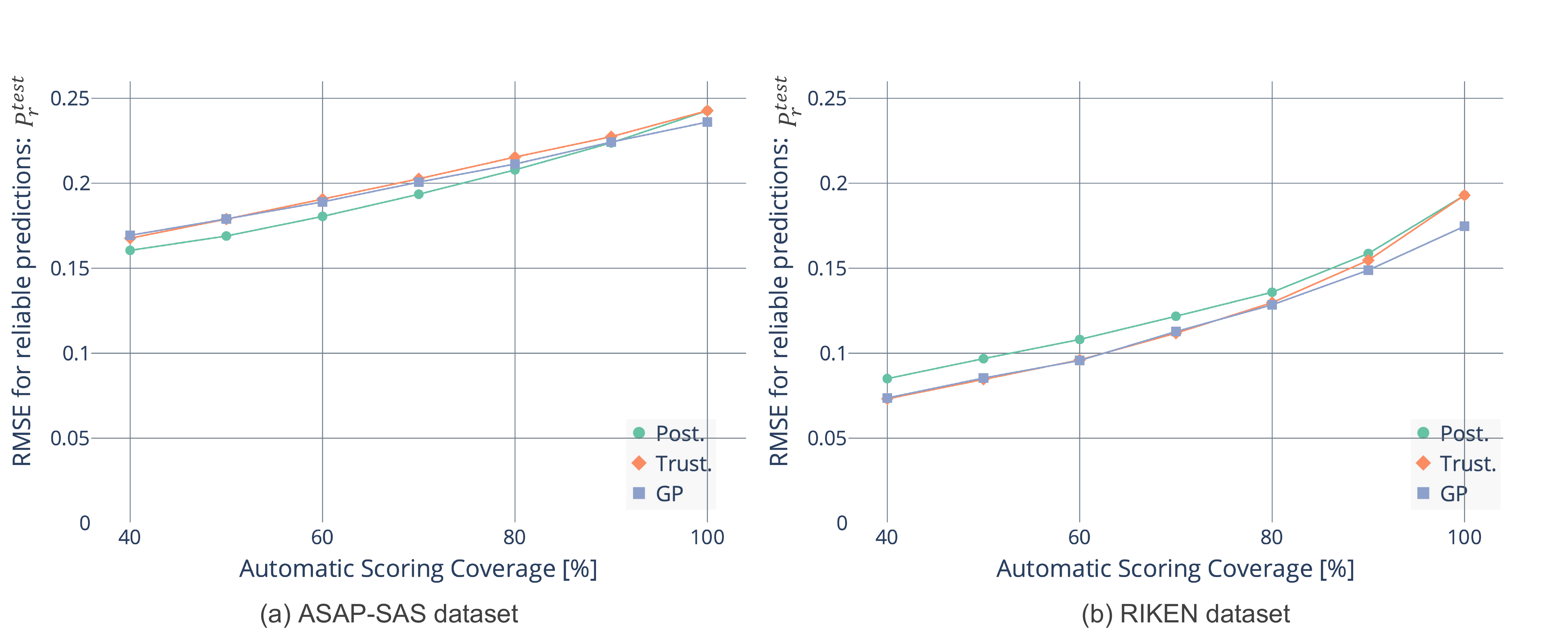}
\caption{Changes of RMSE when the prediction with the top n\% confidence score is adopted as automatic scoring result in the test set. Green markers represent posterior (Post.), red ones the trust score (Trust.), and blue ones, the Gaussian process (GP). The markers represent mean values.}
\label{fig:rmse_coverage_curve}
\end{figure}

\subsection{Setting}
As described in section \ref{ssec:scoring_model}, we used the pretrained BERT~\cite{devlin-etal-2019-bert} as the encoder for the automatic scoring model and use the vectors of CLS tokens as feature vectors for predicting answers\footnote{We used pretrained BERTs from  \url{https://huggingface.co/bert-base-uncased} for English and  \url{https://github.com/cl-tohoku/bert-japanese} for Japanese}
\footnote{Quadratic weghted kappa (QWK) of
our model is 0.722 for the ASAP-SAS dataset,  which is comparable to previous studies\cite{riordan-etal-2017-investigating,Surya-deep-learning-shortanswer-2019}}.

 The root mean square error (RMSE) is adopted as the function \texttt{Err} representing the scoring error. To stabilize the experimental results using a small development set, we estimated the threshold value using 250 data sets, which were integrated with five development datasets.

\subsection{Results}

\subsubsection{Correlation between confidence scores and scoring accuracy}
First, we investigate the relationship between confidence estimation and scoring quality. 
Fig. \ref{fig:rmse_coverage_curve} shows changes in RMSE for the top n\% confident predictions in the test set using the ASAP-SAS dataset and the RIKEN dataset.
All methods reduced the RMSE for the higher confident predictions, indicating that there is a correlation between confidence and scoring quality.
We also observed that the scoring error is reduced more for the RIKEN dataset than for the ASAP-SAS dataset.

\begin{figure}[t]
\centering
\includegraphics[width=\textwidth]{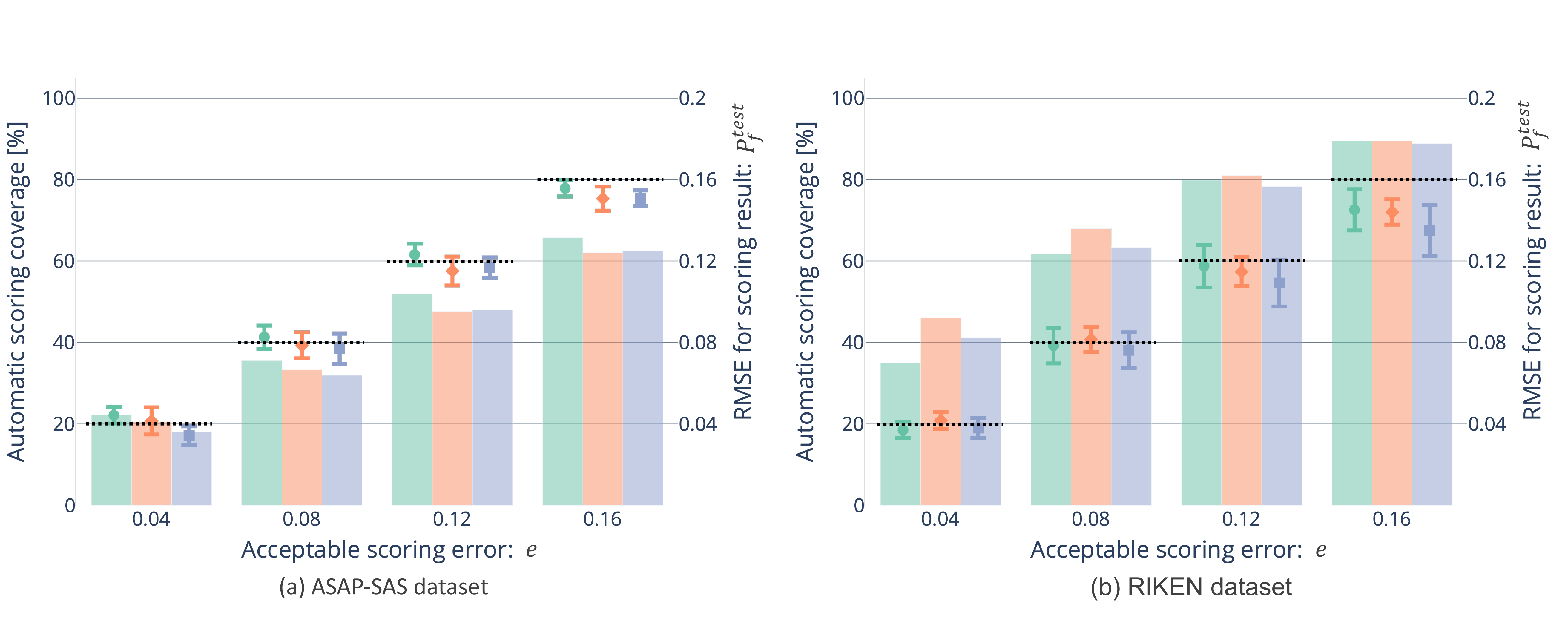}
\caption{Scoring errors for the test data after applying our proposed framework using ASAP-SAS and RIKEN datasets. The bars represent automatic scoring coverage [\%], and the markers and error bars represent RMSE for test set after applying our proposed framework. Green, red, and blue marks represent the results of Post., Trust., and GP, respectively. The error bars represents standard deviation of the RMSE. The dotted line represents the acceptable scoring error $e$. }
\label{fig:main_result}
\end{figure}

\subsubsection{Feasibility of our proposed framework}
Next, we applied the proposed framework to SAS data and confirmed its feasibility.
Fig.~\ref{fig:main_result} shows the percentage of answers automatically scored (automatic scoring coverage [\%]) and the RMSE for the combined scoring result $P^{test}_f$  for human scoring and automatic scoring when the acceptable scoring error was varied from 0.04, 0.08, 0.12, to 0.16 on RMSE.

From Fig.~\ref{fig:main_result}, we can see that we are successfully able to control scoring errors around the acceptable scoring error in most settings for both datasets. 
In terms of the ability to achieve the acceptable scoring error, there is no significant difference in performance between the confidence estimation methods.
The standard deviation indicated by the error bars also shows that our proposed framework performs well with the three confidence estimation methods.
For automatic scoring coverage, the posterior tends to be slightly dominant in ASAP-SAS, whereas the trust score is slightly dominant in the RIKEN dataset.
The slightly higher performance of the trust score for the RIKEN dataset is consistent with Funayama et al~\cite{funayama-etal-2020-preventing}, which was the only study, to the best of our knowledge, to utilize confidence in the SAS field.

\begin{figure}[t]
\centering
\includegraphics[width=\textwidth]{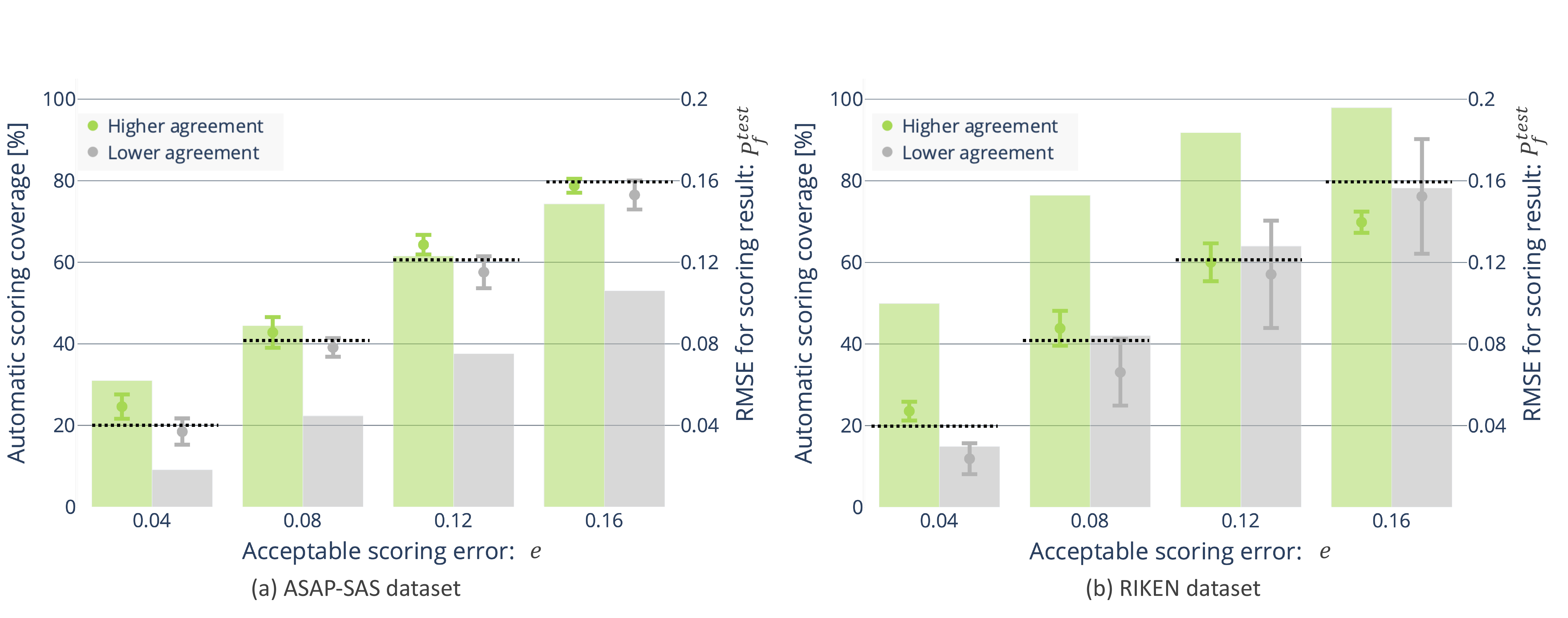}
\caption{The automatic scoring coverage (bars) and  RMSE for $P^{test}_f$ (markers and error bars) when we apply our proposed framework to the two groups: prompts with higher inter grader agreement and prompts with lower inter grader agreement.
}
\label{fig:comparison_based_agreement}
\end{figure}

\subsection{Analysis}
\label{ssec:discussion}
We assumed in Section~\ref{sec:proposed_framework} that professional human graders grade answers perfectly; however, the scoring results between human graders do not perfectly match in the actual dataset.
A previous work~\cite{Williamson_2012}  reported that prompts with higher inter grader agreements between human raters (henceforth IGA) are more suitable for automated scoring in real-life exams, suggesting that IGA can be an important indicator for the actual operation in the application of automatic scoring models.
In this section, we analyze the relevance of the inter grader agreement between two human graders for our framework.

Figure \ref{fig:comparison_based_agreement} shows the results of applying our proposed framework to two groups: one is the group with higher IGAs than the average IGA (green) and the other is the group with lower IGAs (gray).
We use posterior for confidence estimation.\footnote{Posterior is used because there is no significant difference in performance among the three methods and the most widely used way to estimate confidence score is posterior.}
From the results, we observe a large gap of the automatic scoring coverage between the higher IGA and the lower IGA groups, whereas the acceptable scoring errors are generally achieved for both groups.
This indicates that the objective of our framework, which is guaranteeing the grading quality, is achieved regardless of the IGAs of the prompts.
The drawback of a lower IGA is observed in automatic scoring coverage in our framework; the coverage of the higher IGA group is twofold that of the lower group in the strict setting $e=0.04$ for both datasets.

Moreover, we observed that the actual RMSE is smaller than the acceptable scoring error for the prompts in the lower-IGA group, indicating that the framework provides more 'cautious' automatic scoring for this group.
For the noisy training data, an instance with a high prediction confidence tends to be judged as incorrect even though the model predicts the correct score. 
As a result, the scoring quality of the model for predictions with high confidence is underestimated compared with the true scoring quality of the model. 
Then, a strict threshold is selected in the development set, which may lead to such 'cautious' scoring.
Indeed, in the actual situation, when estimating the threshold value in the development set, a human grader is expected to reconfirm and correct the ground truth score for a training instance that has a scoring error despite its high confidence.
Thus, a negative impact of such human scoring errors is expected to be insignificant when using our framework that combines human grading and the confidence of automatic scoring models.

\section{Conclusion}
In recent years, the accuracy of automatic grading systems has considerably improved. Ensuring the scoring quality of SAS models is crucial challenge in the promotion of further applications of SAS models in the education field. 
In this paper, we presented a framework for ensuring the scoring quality of SAS systems by allowing such systems to share the grading task with human graders.
In our experiments, we discovered that the desired scoring quality can be achieved by using our proposed framework.
Our framework is designed to control the overall scoring quality (i.e., RMSE) and is therefore expected to work complementarily with Funayama's work~\cite{funayama-etal-2020-preventing}, which aims to reduce the number of critical scoring errors on individual answers.

\bibliographystyle{splncs04} 
\bibliography{references}

\begin{thebibliography}{10}
\providecommand{\url}[1]{\texttt{#1}}
\providecommand{\urlprefix}{URL }
\providecommand{\doi}[1]{https://doi.org/#1}

\bibitem{Attali_Burstein_2006}
Attali, Y., Burstein, J.: Automated essay scoring with e-rater^^c2^^ae v.2. The
  Journal of Technology, Learning and Assessment  \textbf{4}(3) (Feb 2006)

\bibitem{Burrows2015}
Burrows, S., Gurevych, I., Stein, B.: The eras and trends of automatic short
  answer grading. International Journal of Artificial Intelligence in Education
   \textbf{25}(1),  60--117 (2015)

\bibitem{devlin-etal-2019-bert}
Devlin, J., Chang, M.W., Lee, K., Toutanova, K.: {BERT: Pre-training of Deep
  Bidirectional Transformers for Language Understanding}. In: NAACL-HLT. pp.
  4171--4186 (Jun 2019). \doi{10.18653/v1/N19-1423}

\bibitem{ding-etal-2020-dont}
Ding, Y., Riordan, B., Horbach, A., Cahill, A., Zesch, T.: Don{'}t take
  {``}nswvtnvakgxpm{''} for an answer {--}the surprising vulnerability of
  automatic content scoring systems to adversarial input. In: COLING. pp.
  882--892. International Committee on Computational Linguistics (Dec 2020).
  \doi{10.18653/v1/2020.coling-main.76}

\bibitem{funayama-etal-2020-preventing}
Funayama, H., Sasaki, S., Matsubayashi, Y., Mizumoto, T., Suzuki, J., Mita, M.,
  Inui, K.: Preventing critical scoring errors in short answer scoring with
  confidence estimation. In: ACL-SRW. pp. 237--243. Association for
  Computational Linguistics, Online (Jul 2020).
  \doi{10.18653/v1/2020.acl-srw.32}

\bibitem{gardner2021gpytorch}
Gardner, J.R., Pleiss, G., Bindel, D., Weinberger, K.Q., Wilson, A.G.:
  Gpytorch: Blackbox matrix-matrix gaussian process inference with gpu
  acceleration (2021)

\bibitem{Herwanto-ukara2018}
Herwanto, G., Sari, Y., Prastowo, B., Riasetiawan, M., Bustoni, I.A.,
  Hidayatulloh, I.: Ukara: A fast and simple automatic short answer scoring
  system for bahasa indonesia. In: ICEAP Proceeding Book Vol 2. pp. 48--53 (12
  2018)

\bibitem{horbach-palmer-2016-investigating}
Horbach, A., Palmer, A.: Investigating active learning for short-answer
  scoring. In: BEA. Association for Computational Linguistics, San Diego, CA
  (Jun 2016)

\bibitem{Jang-kass-2014}
Jang, E.S., Kang, S., Noh, E.H., Kim, M.H., Sung, K.H., Seong, T.J.: Kass:
  Korean automatic scoring system for short-answer questions. CSEDU 2014
  \textbf{2},  226--230 (01 2014)

\bibitem{DBLP:conf/nips/JiangKGG18}
Jiang, H., Kim, B., Guan, M.Y., Gupta, M.R.: {To Trust Or Not To Trust A
  Classifier}. In: NIPS. pp. 5546--5557 (2018)

\bibitem{johan-berggren-etal-2019-regression}
Johan~Berggren, S., Rama, T., {\O}vrelid, L.: Regression or classification?
  automated essay scoring for {N}orwegian. In: BEA. pp. 92--102. Association
  for Computational Linguistics, Florence, Italy (Aug 2019).
  \doi{10.18653/v1/W19-4409}

\bibitem{Surya-deep-learning-shortanswer-2019}
Krishnamurthy, S., Gayakwad, E., Kailasanathan, N.: Deep learning for short
  answer scoring. International Journal of Recent Technology and Engineering
  \textbf{7},  1712--1715 (03 2019)

\bibitem{Singla2019GetIS}
Kumar, Y., Aggarwal, S., Mahata, D., Shah, R.R., Kumaraguru, P., Zimmermann,
  R.: Get it scored using autosas ― an automated system for scoring short
  answers. In: AAAI/IAAI/EAAI. AAAI Press (2019).
  \doi{10.1609/aaai.v33i01.33019662}

\bibitem{Leacock}
Leacock, C., Chodorow, M.: {C-rater: Automated Scoring of Short-Answer
  Questions}. Computers and the Humanities  \textbf{37}(4),  389--405 (2003),
  \url{https://doi.org/10.1023/A:1025779619903}

\bibitem{mizumoto-etal-2019-analytic}
Mizumoto, T., Ouchi, H., Isobe, Y., Reisert, P., Nagata, R., Sekine, S., Inui,
  K.: {Analytic Score Prediction and Justification Identification in Automated
  Short Answer Scoring}. In: BEA. pp. 316--325 (2019).
  \doi{10.18653/v1/W19-4433}

\bibitem{Rasmussen2004}
Rasmussen, C.E.: Gaussian processes in machine learning. In: Advanced Lectures
  on Machine Learning: ML Summer Schools 2003. pp. 63--71 (2004)

\bibitem{riordan-etal-2017-investigating}
Riordan, B., Horbach, A., Cahill, A., Zesch, T., Lee, C.M.: {Investigating
  neural architectures for short answer scoring}. In: BEA. pp. 159--168 (2017).
  \doi{10.18653/v1/W17-5017}

\bibitem{sychev2020automatic}
Sychev, O., Anikin, A., Prokudin, A.: Automatic grading and hinting in
  open-ended text questions. Cognitive Systems Research  \textbf{59},  264--272
  (2020)

\bibitem{wang-etal-2019-inject}
Wang, T., Inoue, N., Ouchi, H., Mizumoto, T., Inui, K.: {Inject Rubrics into
  Short Answer Grading System}. In: DeepLo. pp. 175--182 (Nov 2019).
  \doi{10.18653/v1/D19-6119}

\bibitem{Williamson_2012}
Williamson, D.M., Xi, X., Breyer, F.J.: A framework for evaluation and use of
  automated scoring. Educational Measurement: Issues and Practice
  \textbf{31}(1),  2--13 (2012). \doi{10.1111/j.1745-3992.2011.00223.x}

\bibitem{Woods_2017}
Woods, B., Adamson, D., Miel, S., Mayfield, E.: Formative essay feedback using
  predictive scoring models. p. 2071^^e2^^80^^932080. KDD '17, Association for
  Computing Machinery (2017). \doi{10.1145/3097983.3098160}

\end{thebibliography}

\end{document}